\documentclass[10pt,twocolumn,letterpaper]{article}

\usepackage{iccv}
\usepackage{times}
\usepackage{epsfig}
\usepackage{graphicx}
\usepackage{amsmath}
\usepackage{amssymb}

\usepackage{subfigure}
\usepackage{multirow} 
\usepackage{upgreek} 

\usepackage[breaklinks=true,colorlinks,bookmarks=false]{hyperref}

\iccvfinalcopy 

\def\iccvPaperID{1556} 

\ificcvfinal\fi
\begin{document}

\title{Deeply-Learned Part-Aligned Representations for Person Re-Identification}

\author{
	Liming Zhao$^\dag$
	\quad
	Xi Li$^\dag$\thanks{Corresponding authors.}
	\quad
	Jingdong Wang$\ddag$$^*$
	\quad
	Yueting Zhuang$^\dag$
	\\
	$^\dag$Zhejiang University
	\quad
	$^\ddag$Microsoft Research
	\\
	{\tt\small \{zhaoliming,xilizju,yzhuang\}@zju.edu.cn}
	\quad
	{\tt\small jingdw@microsoft.com}
}

\maketitle
\thispagestyle{empty}


\begin{abstract}
In this paper,
we address the problem of person re-identification,
which refers to associating the persons
captured from different cameras.
We propose a simple yet effective human part-aligned representation
for handling the body part misalignment problem.
Our approach decomposes the human body into regions (parts)
which are discriminative for person matching,
accordingly computes the representations over the regions,
and aggregates
the similarities computed between the corresponding regions
of a pair of probe and gallery images
as the overall matching score.
Our formulation,
inspired by attention models,
is a deep neural network
modeling the three steps together,
which is learnt
through minimizing the triplet loss function
without requiring body part labeling information.
Unlike most existing deep learning algorithms
that learn a global or spatial partition-based local representation,
our approach performs human body partition,
and thus is more robust to pose changes
and various human spatial distributions in the person bounding box.
Our approach shows state-of-the-art results
over standard datasets,
Market-$1501$,
CUHK$03$, CUHK$01$
and VIPeR.
\footnote{This work was done when Liming Zhao was an intern at Microsoft Research.}
\end{abstract}

\section{Introduction}
Person re-identification
is a problem of
associating the persons
captured
from different cameras located
at different physical sites.
If the camera views are overlapped,
the solution is trivial:
the temporal information is reliable
to solve the problem.
In some real cases,
the camera views are significantly disjoint
and the temporal transition time between cameras varies greatly,
making the temporal information not enough
to solve the problem,
and thus this problem becomes more challenging.
Therefore, a lot of solutions exploiting various cues,
such as appearance~\cite{Farenzena2010symmetry,mignon2012pcca,deepreid2014,LOMO2015},
which is also the interest in this paper,
have been developed.

\begin{figure}[t]
   \centering
   \small
   \includegraphics[width=1\linewidth]{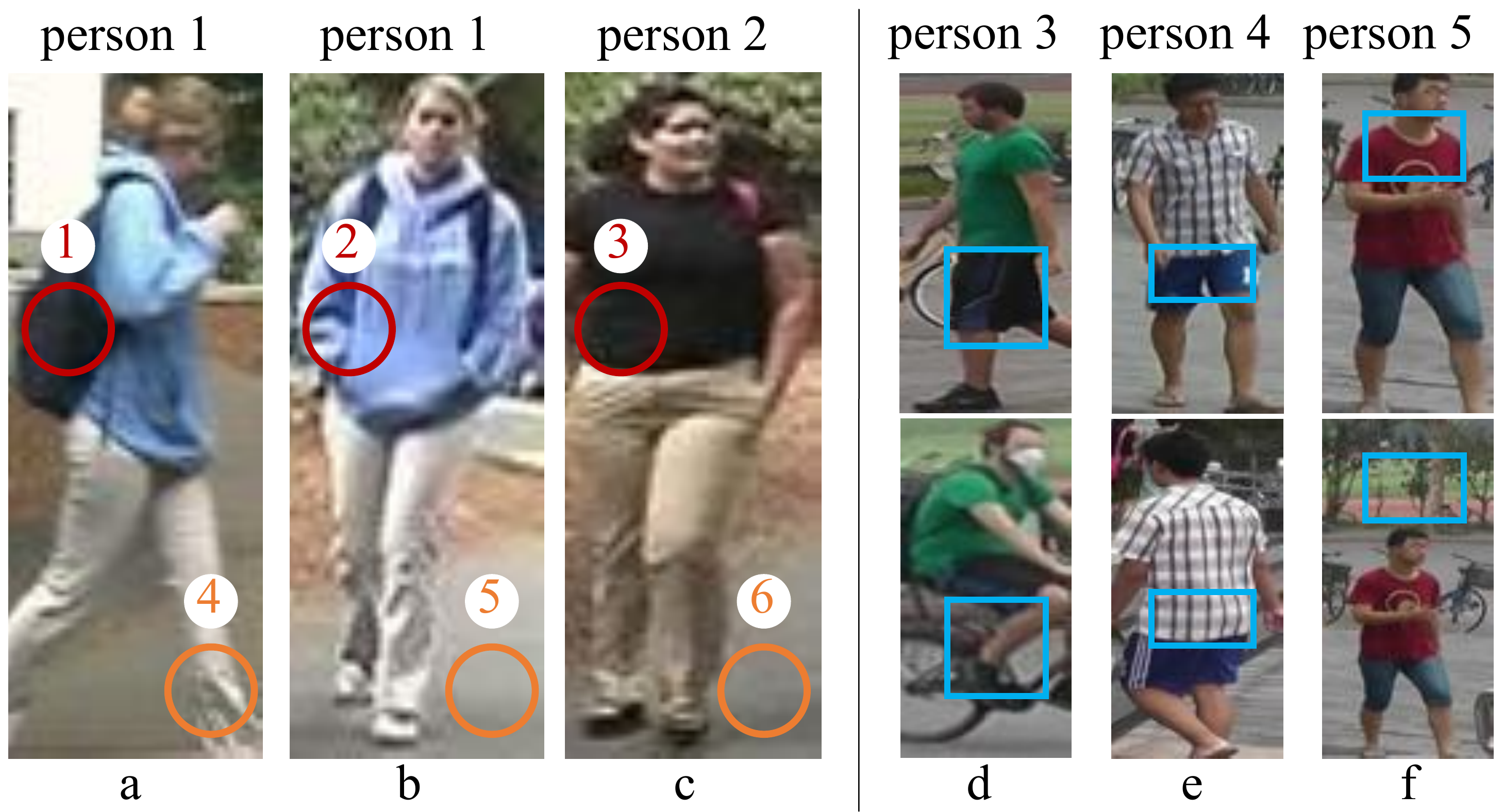}
   \caption{Illustrating the necessity
       of body part partition (best viewed in color).
       Using spatial partition
       without further processing,
       the regions $(1)$ and $(2)$, as well as $(4)$ and $(5)$, are not matched though they are from the same person;
       but the regions $(1)$ and $(3)$, as well as $(5)$ and $(6)$, which are from different persons, are matched.
       With body part decomposition,
       there is no such mismatch.
       More examples are shown in d, e, and f.   }
   \label{fig:misalignment}
   \vspace{-.3cm}
\end{figure}

Recently, deep neural networks have
been becoming a dominate solution
for the appearance representation.
The straightforward way is to extract
a global representation~\cite{PaisitkriangkraiSV15,wu2016enhanced,chen2016deep},
using the deep network
pretrained over ImageNet
and optionally fine-tuned
over the person re-identification dataset.
Local representations
are computed typically by partitioning the person bounding box
into cells,
e.g.,
dividing the images into horizontal stripes~\cite{deepmetric2014, ChengGZWZ16,VariorSLXW16}
or grids~\cite{deepreid2014, improved2015},
and extracting deep features over the cells.
These solutions are based on the assumption
that the human poses and
the spatial distributions of the human body
in the bounding box
are similar.
In real cases,
for example,
the bounding box is detected rather
than manually labeled
and thus the human may be at different positions,
or the human poses are different,
such an assumption does not hold.
In other words,
spatial partition is not well aligned with human body parts.
Thus, person re-identification,
even with subsequent complex matching techniques
(e.g.,~\cite{improved2015, deepreid2014}) to eliminate the misalignment,
is often not quite reliable.
Figure~\ref{fig:misalignment} provides illustrative examples.


In this paper,
we propose a part-aligned human representation,
which addresses the above problem
instead in the representation learning stage.
The key idea is straightforward:
detect the human body regions that are discriminative for person matching,
compute the representations over the parts,
and then aggregate the similarities
that are computed between the corresponding parts.
Inspired by attention models~\cite{XuBKCCSZB15},
we present a deep neural network method,
which jointly models body part extraction and
representation computation, and learns model parameters
through maximizing the re-identification quality
in an end-to-end manner,
without requiring the labeling information
about human body parts.
In contrast to spatial partition,
our approach performs human body part partition,
thus is more robust to human pose changes
and various human spatial distributions in the bounding box.
Empirical results demonstrate
that our approach
achieves competitive/superior performance
over standard datasets:
Market-$1501$,
CUHK$03$, CUHK$01$
and VIPeR.

%

\section{Related Work}
There are two main issues
in person re-identification:
representation and matching.
Various solutions, separately or jointly
addressing the two issues, have been developed.

\vspace{0.1cm}
\noindent\textbf{Separate solutions.}
Various hand-crafted representations
have been developed,
such as the ensemble of local features (ELF)~\cite{viewpoint2008}, fisher vectors (LDFV)~\cite{ma2012local}, local maximal occurrence representation (LOMO)~\cite{LOMO2015},
hierarchal Gaussian descriptor (GOG)~\cite{MatsukawaOSS16},
and so on.
Most of the representations are designed with the goal of handling light variance,
pose/view changes, and so on.
Person attributes or salient patterns,
such as female/male, wearing hat or not,
have also been exploited
to distinguish persons~\cite{SuYZTDLG15, SuZX0T16, ZhaoOW14}.

A lot of similarity/metric learning techniques~\cite{null2016,ZhangLLIR16, PaisitkriangkraiSV15, psd2015, KodirovXFG16}
have been applied or designed
to learn metrics,
robust to light/view/pose changes, for person matching.
The recent developments include
soft and probabilistic patch matching for handling pose misalignment~\cite{dapeng2015, ChenYCZ16, correspondence2015},
similarity learning for dealing with probe and gallery images
with different resolutions~\cite{LiZWXG15, JingZWYKYHX15},
connection with transfer learning~\cite{PengXWPGHT16,ShiHX15},
reranking inspired by the connection with image search~\cite{ZhengWTHLT15, GarciaMMG15},
partial person matching~\cite{ZhengLXLLG15},
human-in-the-loop learning~\cite{MartinelDMR16,WangGZX16},
and so on.

%
%
%

\vspace{0.1cm}
\noindent\textbf{Deep learning-based solutions.}
The success of deep learning in image classification
has been inspiring a lot of studies
in person re-identification.
The off-the-shelf CNN features,
extracted from the model trained over ImageNet,
without fine tuning,
does not show the performance gain~\cite{PaisitkriangkraiSV15}.
The promising direction is
to learn the representation
and the similarity jointly,
except some works~\cite{XiaoLOW16, ZhengBSWSWT16}
that do not learn the similarity
but adopt the classification loss
by regarding the images about one person
as a category.

The network typically consists of two subnetworks:
one for feature extraction
and the other for matching.
The feature extraction subnetwork
could be simply
(\romannumeral1) a shallow network~\cite{deepreid2014}
with one or two convolutional and max-pooling layers
for feature extraction,
or
(\romannumeral2) a deep network,
e.g., VGGNet and its variants~\cite{vggnet2014, PersonNet2016}
and GoogLeNet~\cite{googlenet2015, DCSL2016ijcai},
which are pretrained over ImageNet
and fine-tuned for person re-identification.
The feature representation can be (\romannumeral1) a global feature,
e.g., the output of the fully-connected layer
~\cite{chen2016deep,xiaoli2016end},
which does not explicitly model the spatial information,
or
(\romannumeral2) a combination
(e.g., concatenation~\cite{deepmetric2014, ChengGZWZ16}
or contextual fusion~\cite{VariorSLXW16}) of the features
over regions,
e.g., horizontal stripes~\cite{deepmetric2014, ChengGZWZ16,VariorSLXW16},
or grid cells~\cite{deepreid2014, improved2015},
which are favorable for the later matching process
to handle body part misalignment.
Besides,
the cross-dataset information~\cite{XiaoLOW16} is also exploited
to learn an effective representation.

The matching subnetwork can simply be a loss layer
that penalizes the misalignment
between learnt similarities and
ground-truth similarities,
e.g., pairwise loss~\cite{deepmetric2014, VariorSLXW16, deepreid2014, improved2015,ShiYZLLZL16},
triplet loss and its variants~\cite{tripletdeepreid2015,ChengGZWZ16,SuZX0T16,WangZLZZ16}.
Besides using
the off-the-shelf similarity function~\cite{deepmetric2014, VariorSLXW16,ChengGZWZ16},
e.g., cosine similarity or Euclidean distance,
for comparing the feature representation,
specific matching schemes
are designed
to eliminate the influence
from body part misalignment.
For instance,
a matching subnetwork
conducts convolution and max pooling operations,
over the differences~\cite{improved2015}
or the concatenation~\cite{deepreid2014,DCSL2016ijcai}
of the representations over grid cells
of a pair of person images,
to handle the misalignment problem.
The approach with so called single-image and cross-image
representations~\cite{WangZLZZ16} essentially
combines the off-the-shelf distance
and the matching network handling the misalignment.
Instead of only matching the images
over the final representation,
the matching map in the intermediate features is used
to guide the feature extraction in the later layers
through a gated CNN~\cite{VariorHW16}.

\vspace{0.1cm}
\noindent\textbf{Our approach.}
In this paper, we focus on the feature extraction part
and introduce a human body part-aligned representation.
Our approach is related to but different from
the previous part-aligned approaches
(e.g., part/pose detection~\cite{ChengCSBM11, XuLZL13, BakCBT10a, ZhengHLY17}),
which need to train a part/pose segmentation or detection model
from the labeled part mask/box or pose ground-truth
and subsequently extract representations,
where the processes are conducted separately.
In contrast,
our approach does not require those labeling information,
but only uses the similarity information
(a pair of person images are about the same person
or different persons),
to learn the part model for person matching.
The learnt parts are different from the conventional human body parts,
e.g., Pascal-Person-Parts~\cite{pascalpart},
and are specifically for person matching,
implying that our approach potentially performs better,
which is verified by
empirical comparisons with
the algorithms based on the state-of-the-art part segmentation approach (deeplab~\cite{deeplabv1})
and
pose estimator (convolutional pose machine~\cite{WeiRKS16}).

Our human body part estimation scheme is inspired
by the attention model that is successfully applied to
many applications such as image captioning~\cite{XuBKCCSZB15}.
Compared to the work~\cite{LiuFQJY16}
that is based on attention models and LSTM,
our approach is simple and easily implemented,
and empirical results show that our approach performs better.


\section{Our Approach}

%

Person re-identification aims to
find the images
that are about the same identity
with the probe image
from a set of gallery images.
It is often regarded
as a ranking problem:
given a probe image,
the gallery images about the same identity
are thought closer to the probe image
than the gallery images about different identities.

The training data is typically
given as follows.
Given a set of images
$\mathcal{I} =\{\mathbf{I}_1, \mathbf{I}_2, \dots, \mathbf{I}_N\}$,
we form the training set
as a set of triplets,
$\mathcal{T} = \{(\mathbf{I}_i, \mathbf{I}_j, \mathbf{I}_k)\}$,
where
$(\mathbf{I}_i, \mathbf{I}_j)$ is a positive pair of images
that are about the same person
and $(\mathbf{I}_i, \mathbf{I}_k)$ is a negative pair of images
that are about different persons.

Our approach formulates the ranking problem
using the triplet loss function,
\begin{align}
&\ell_{\operatorname{triplet}}(\mathbf{I}_i, \mathbf{I}_j, \mathbf{I}_k) \nonumber \\
=~& [d(h(\mathbf{I}_i), h(\mathbf{I}_j)) - d(h(\mathbf{I}_i), h(\mathbf{I}_k)) + m]_{+}.
\end{align}
Here $(\mathbf{I}_i, \mathbf{I}_j, \mathbf{I}_k) \in \mathcal{T}$.
$m$ is the margin
by which the distance between a negative pair of images
is greater than
that between a positive pair of images.
In our implementation,
$m$ is set to $0.2$ similar to~\cite{facenet2015}.
$d(\mathbf{x}, \mathbf{y})=\|\mathbf{x} - \mathbf{y}\|_2^2$ is a Euclidean distance.
$[z]_+ = \max(z, 0)$ is the hinge loss.
$h(\mathbf{I})$ is a feature extraction network
that extracts the representation of the image $\mathbf{I}$
and will be discussed in detail later.
The whole loss function is as follows,
\begin{align}
\mathcal{L}(h)
= \frac{1}{|\mathcal{T}|}\sum_{(\mathbf{I}_i, \mathbf{I}_j, \mathbf{I}_k) \in \mathcal{T}} \ell_{\operatorname{triplet}}(\mathbf{I}_i, \mathbf{I}_j, \mathbf{I}_k),
\label{eqn:wholeloss}
\end{align}
where $|\mathcal{T}|$ is the number of triplets
in $\mathcal{T}$.

\subsection{Part-Aligned Representation}
The part-aligned representation extractor,
is a deep neural network,
consisting of a fully convolutional neural network (FCN)
whose output is an image feature map,
followed by a part net
which detects part maps
and outputs the part features
extracted over the parts.
Rather than partitioning the image box spatially
to grid cells or horizontal stripes,
our approach aims to
partition the human body
to aligned parts.

\begin{figure}[t]
\centering
        \includegraphics[width=1\linewidth]{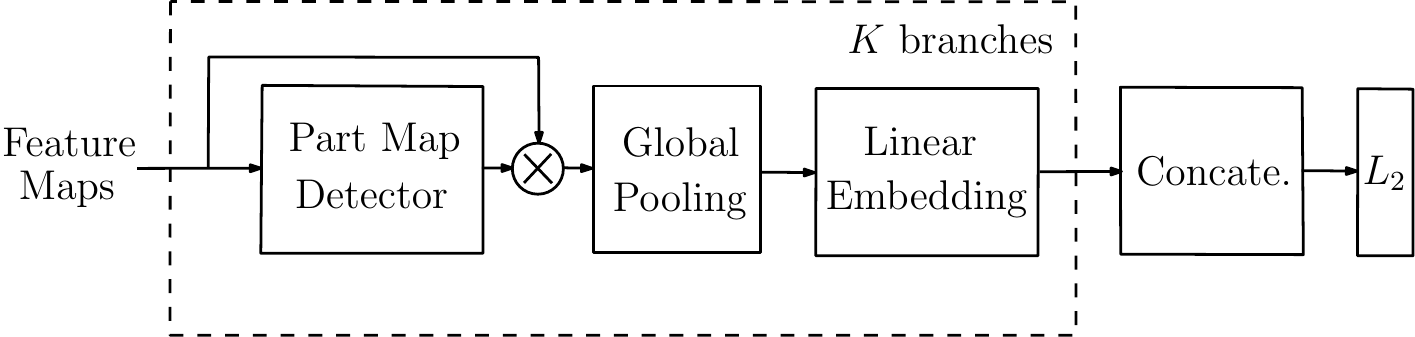}
        \caption{Illustrating the part net.
    It consists of $K$ branches.
    Each branch takes the image feature map as the input and estimates a part map,
    which is used for weighting
    the image feature map followed by an average pooling operator.
    The part features from the $K$ branches are concatenated
    as the final human representation.
}
\label{fig:partnet}
\vspace{-1.5em}
\end{figure}

The part net,
as illustrated in Figure~\ref{fig:partnet},
contains several branches.
Each branch receives
the image feature map from the FCN
as the input,
detects a discriminative region (part\footnote{In this paper,
we use the two terms,
part and region,
interchangeably for
the same meaning.}),
and extracts the feature over the detected region
as the output.
As we will see, the detected region usually
lies in the human body region,
which is as expected
because these regions are informative for person matching.
Thus, we call the net as a part net.
Let a $3$-dimensional tensor
$\mathbf{T}$
represent the image feature maps computed from the FCN
and thus $t(x, y, c)$
represent the $c$th response
over the location $(x,y)$.
The part map detector estimates
a $2$-dimensional map $\mathbf{M}_k$,
where $m_k(x,y)$ indicates
the degree that the location $(x,y)$
lies in the $k$th region,
from the image feature map $\mathbf{T}$:
\begin{align}
\mathbf{M}_k = N_{\operatorname{MapDetector_k}}(\mathbf{T}),
\label{eqn:partdetector}
\end{align}
where$N_{\operatorname{MapDetector_k}}(\cdot)$ is a region map detector
implemented as a convolutional network.

The part feature map $\mathbf{T}_k$
for the $k$th region
is computed through a weighting scheme,
\begin{align}
t_k(x, y, c) = t(x, y, c) \times m_k(x, y),
\end{align}
followed by an average pooling operator,
$\bar{\mathbf{f}}_k = \operatorname{AvePooling}(\mathbf{T}_k)$,
where $\bar{f}_k(c) = \operatorname{Average}_{x,y}[t_k(x, y, c)]$.
Then a linear dimension-reduction layer,
implemented as a fully-connected layer,
is performed
to reduce $\bar{\mathbf{f}}_k$
to a $d$-dimensional feature vector
$\mathbf{f}_k = \mathbf{W}_{FC_k} \bar{\mathbf{f}}_k$.
Finally,
we concatenate all the part features,
\begin{align}
\mathbf{f} = [\mathbf{f}_1^{\top}~\mathbf{f}_2^{\top}~\dots~\mathbf{f}_K^{\top}]^{\top},
\end{align}
and perform an $L_2$ normalization,
yielding the human representation $h(\mathbf{I})$.

\subsection{Optimization}
We learn the network parameters,
denoted by $\boldsymbol{\uptheta}$,
by minimizing the summation of
triplet loss functions
over triplets
formulated in Equation~\ref{eqn:wholeloss}.
The gradient is computed as
\begin{align}
\frac{\partial \mathcal{L}}{\partial \boldsymbol{\uptheta}}
= \frac{1}{|\mathcal{T}|} \sum_{(\mathbf{I}_i, \mathbf{I}_j, \mathbf{I}_k) \in \mathcal{T}}
\frac{\partial \ell_{\operatorname{triplet}} (\mathbf{I}_i, \mathbf{I}_j, \mathbf{I}_k) }{\partial \boldsymbol{\uptheta}}.
\end{align}
We have\footnote{The gradient at the non-differentiable point is omitted
like the common way to handle this case in deep learning.}
\begin{align}
&\frac{\partial \ell_{\operatorname{triplet}} (\mathbf{I}_i, \mathbf{I}_j, \mathbf{I}_k) }{\partial \boldsymbol{\uptheta}} \nonumber \\
=~&
\delta_{\ell_{\operatorname{triplet}} (\mathbf{I}_i, \mathbf{I}_j, \mathbf{I}_k) > 0} \times
2[\frac{\partial h(\mathbf{I}_i)}{\partial \boldsymbol{\uptheta}} (h(\mathbf{I}_k) - h(\mathbf{I}_j)) + \nonumber \\
& \frac{\partial h(\mathbf{I}_j)}{\partial \boldsymbol{\uptheta}}(h(\mathbf{I}_j) - h(\mathbf{I}_i)) + \frac{\partial h(\mathbf{I}_k)}{\partial \boldsymbol{\uptheta}}(h(\mathbf{I}_i) - h(\mathbf{I}_k))]. \nonumber
\end{align}
Thus,
we transform the gradient to the following form,
\begin{align}
\frac{\partial \mathcal{L}}{\partial \boldsymbol{\uptheta}}
= \frac{1}{|\mathcal{T}|}\sum_{n=1}^N \frac{\partial h(\mathbf{I}_n)}{\partial \boldsymbol{\uptheta}} \boldsymbol{\upalpha}_n,
\label{eqn:weightedgradient}
\end{align}
where $\boldsymbol{\upalpha}_n$ is a weight vector
depending on the current network parameters,
and computed as follows,
\begin{align}\vspace{-.1cm}
~&\boldsymbol{\upalpha}_n = 2[\sum_{(\mathbf{I}_n, \mathbf{I}_j, \mathbf{I}_k) \in \mathcal{T}}
\delta_{\ell_{\operatorname{triplet}} (\mathbf{I}_n, \mathbf{I}_j, \mathbf{I}_k) > 0}
(h(\mathbf{I}_k) - h(\mathbf{I}_j))
+ \nonumber \\
~& \sum_{(\mathbf{I}_i, \mathbf{I}_n, \mathbf{I}_k) \in \mathcal{T}}
\delta_{\ell_{\operatorname{triplet}} (\mathbf{I}_i, \mathbf{I}_n, \mathbf{I}_k) > 0}
(h(\mathbf{I}_n) - h(\mathbf{I}_i))
+ \nonumber  \\
~& \sum_{(\mathbf{I}_i, \mathbf{I}_j, \mathbf{I}_n) \in \mathcal{T}}
\delta_{\ell_{\operatorname{triplet}} (\mathbf{I}_i, \mathbf{I}_j, \mathbf{I}_n) > 0}
(h(\mathbf{I}_i) - h(\mathbf{I}_n))].
\label{eqn:gradientweight}
\vspace{-.1cm}
\end{align}
Equation~\ref{eqn:weightedgradient}
suggests that
the gradient for the triplet loss
is computed like that for the unary classification loss.
Thus,
in each iteration of SGD (stochastic gradient descent)
we can draw a mini-batch of ($M$) samples rather than
sample a subset of triplets:
one pass of forward propagation
to compute the representation $h(\mathbf{I}_n)$ of each sample,
compute the weight $\boldsymbol{\upalpha}_n$
over the mini-batch,
compute the gradient $\frac{\partial h(\mathbf{I}_n)}{\boldsymbol{\uptheta}}$,
and finally aggregate the gradients over the mini-batch of samples.
Directly drawing a set of triplets
usually leads to that a larger number of (more than $M$) samples are contained
and thus the computation is more expensive than our mini-batch sampling scheme.

\subsection{Implementation details}
\label{sec:implementationdetails}
\noindent\textbf{Network architecture.}
We use a sub-network of the first version of GoogLeNet~\cite{googlenet2015},
from the image input
to the output of \textit{inception\_{4e}},
followed by a $1 \times 1$ convolutional layer
with the output of $512$ channels,
as the image feature map extraction network.
Specifically,
the person image box is resized
to $160 \times 80$
as the input,
and thus the size of the feature map of the feature map extraction network
is $10 \times 5$
with $512$ channels.
For data preprocessing, we use the standard horizontal flips of the resized image.
In the part net, the part estimator
($N_{\operatorname{MapDetector_k}}$ in Equation~\ref{eqn:partdetector})
is simply a $1 \times 1$ convolutional layer
followed by a nonlinear sigmoid layer.
There are $K$ part detectors,
where $K$ is determined by cross-validation and
empirically studied in Section~\ref{sec:empiricalstudy}.

\vspace{0.1cm}
\noindent\textbf{Network Training.}
We use the stochastic gradient descent algorithm to
train the whole network based on Caffe~\cite{jia2014caffe}.
The image feature map extraction part is initialized
using the GoogLeNet model, pretrained over ImageNet.
In each iteration,
we sample a mini-batch of $400$ images,
e.g., there are on average $40$ identities
with each containing $10$ images
on Market-$1501$ and CUHK$03$.
In total,
there are about $1.4$ million triplets
in each iteration.
From Equation~\ref{eqn:gradientweight},
we see that only a subset of triplets,
whose predicted similarity order
is not consistent to the ground-truth order,
i.e., $\ell_{\operatorname{triplet}} (\mathbf{I}_n, \mathbf{I}_j, \mathbf{I}_k) > 0$,
are counted for the weight ($\boldsymbol{\uptheta}$) update,
and accordingly we use the number of counted triplets
to replace $|\mathcal{T}|$ in Equation~\ref{eqn:weightedgradient}.

We adopt the initial learning rate, $0.01$,
and divide it by $5$ every $20K$ iterations.
The weight decay is $0.0002$
and the momentum for gradient update is $0.9$.
Each model is trained for $50K$ iterations within around $12$ hours on a K40 GPU.
For testing, it takes on average $0.005$ second on one GPU
to extract the part-aligned representation.

\begin{figure}[t]
\centering
    \includegraphics[width=1\linewidth]{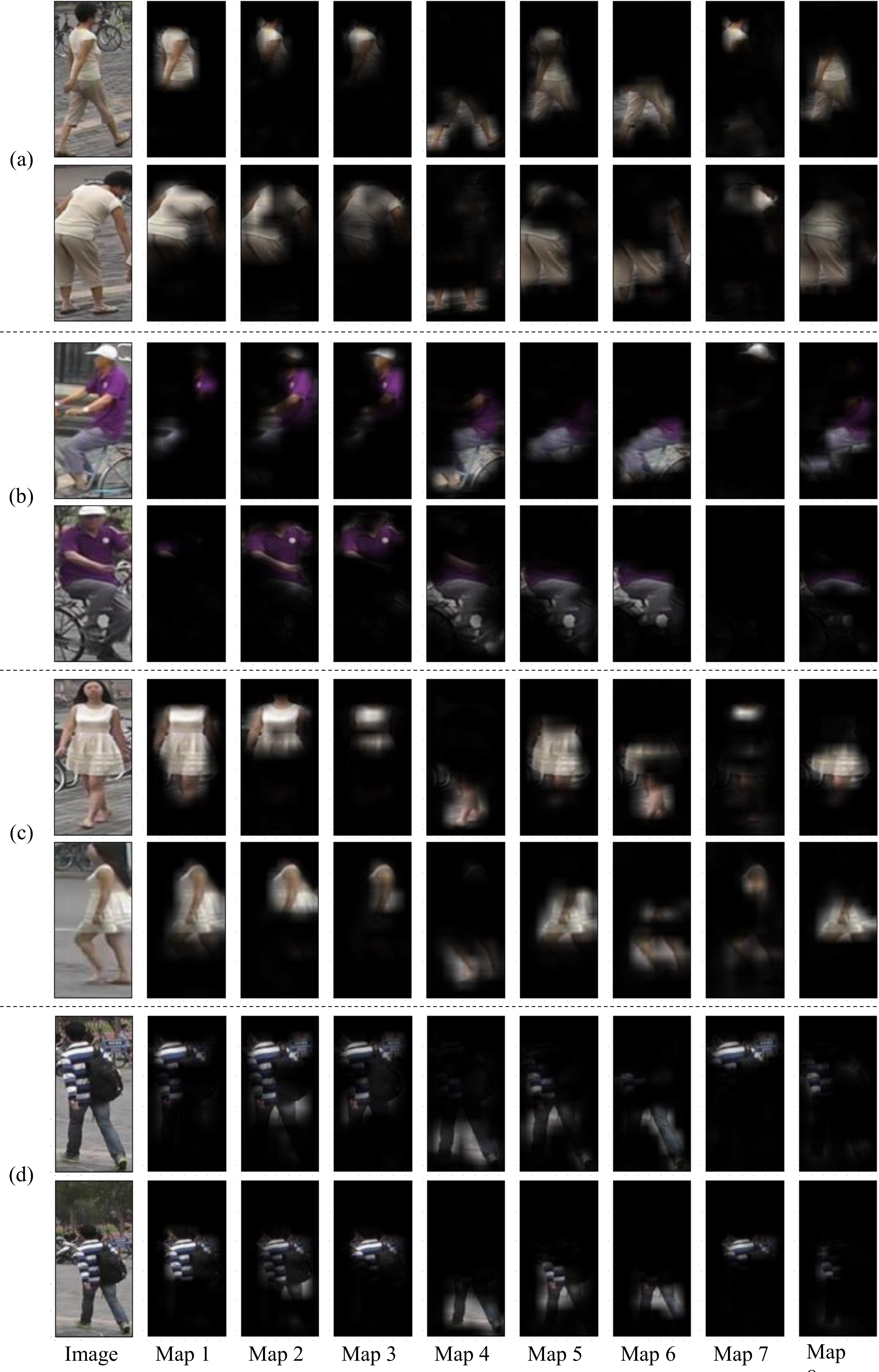}
\caption{Examples of the part maps learnt by the part map estimator
    for test images (best viewed in color).}
\label{fig:weighting_maps}
\vspace{-1.5em}
\end{figure}

\subsection{Discussions}
\noindent\textbf{Body part partition and spatial partition.}
Spatial partition, e.g.,
grid or stride-based,
may not be well aligned with human body parts,
due to pose changes or various human spatial distributions in the human image box.
Thus, matching techniques, e.g., through complex networks~\cite{improved2015, deepreid2014, DCSL2016ijcai},
have been developed
to eliminate the misalignment problem.
In contrast,
our approach addresses this problem
in the representation stage,
with a simple Euclidean distance
for person matching,
which potentially makes existing fast similarity search algorithms
easily applied,
and thus the online search stage more efficient. %

Figure~\ref{fig:weighting_maps} shows the examples
about the parts our approach learns
for the test images.
It can be seen that the parts
are generally well aligned for the pair of images about the same person:
the parts almost describe the same human body regions,
except that one or two parts in the pair of images describe
different regions, e.g.,
the first part in Figure~\ref{fig:weighting_maps} (b).
In particular,
the alignment is also good for the examples of Figure~\ref{fig:weighting_maps} (c, d),
where the person
in the second image is spatially distributed
very differently
from the person in the first image:
one is on the right in Figure~\ref{fig:weighting_maps} (c),
and one is small and on the bottom in Figure~\ref{fig:weighting_maps} (d).

In addition,
we empirically compare our approach with two spatial partition based methods:
dividing the image box into $5$ horizontal stripes or $5\times5$ girds
to form region maps.
We use the region maps to replace the part mask in our approach
and then learn the spatial partition-based representation.
The results shown in Table~\ref{tab:spatialpartition}
demonstrate that the human body part partition method is more effective.

\begin{table}[htbp]
  \centering
  \caption{The performance ($\%$) of our approach and spatial partition based methods
  (stripe and grid)
  over Market-$1501$ and CUHK$03$.
  }
  \label{tab:spatialpartition}%
  \footnotesize{
    \begin{tabular}{c|c|cccc}
  \hline
  Dataset & Method & rank-$1$ & rank-$5$ & rank-$10$ & rank-$20$ \\
  \hline
  & ours  & $\mathbf{81.0}$  & $\mathbf{92.0}$  & $\mathbf{94.7}$  & $\mathbf{96.4}$  \\
  Market-$1501$ & stripe &  $74.1$     &    $89.0$    &   $92.3$     &  $95.1$ \\
  & grid  &   $73.4$    &   $88.2$    &  $91.8$     &  $94.4$\\
  \hline
  & ours  & $\mathbf{85.4}$  & $\mathbf{97.6}$  & $\mathbf{99.4}$  & $\mathbf{99.9}$  \\
  CUHK$03$ & stripe & $81.4$  & $97.1$  & $99.3$  & $99.7$  \\
  & grid  & $78.2$  & $96.7$  & $99.2$  & $99.8$  \\
  \hline
    \end{tabular}%
  }
  \vspace{-1.5em}
\end{table}%

\vspace{0.1cm}
\noindent\textbf{Learnt body parts.}
We have several observations
about the learnt parts.
The head region
is not included.
This is because the face is not frontal
and with low resolution
and accordingly not reliable
for differentiating different persons.
The skin regions are often also not included
except the arms located nearby the top body in Figure~\ref{fig:weighting_maps} (c)
as the skin does not provide discriminant information,
e.g., the leg skins in Figure~\ref{fig:weighting_maps} (c)
are not included
while the legs with trousers in Figure~\ref{fig:weighting_maps} (b)
are included
in Map$4$-$6$ .

From Figure~\ref{fig:weighting_maps},
we can see that the first three maps,
Map$1$ - Map$3$,
are about the top clothing.
There might be some redundancy.
In the examples of Figure~\ref{fig:weighting_maps} (c,d),
the first two masks are very close.
In contrast,
in the examples of Figure~\ref{fig:weighting_maps} (b),
the masks are different,
and are different regions of the top,
though all are about the top clothing.
In this sense,
the first three masks act like
a mixture model
to describe the top clothing
as the top parts are various
due to pose and view variation.
Similarly, Map$4$ and Map$6$ are both about the bottom.

\begin{table}
\centering
\caption{
The performance of our approach, and separate part segmentation
over Market-$1501$ and CUHK-$03$.}
\label{tab:SeparatePartSegmentation}
\resizebox{1\linewidth}{!}{
\begin{tabular}{c|c|c c c c}
  \hline
  Dataset & Method & rank-$1$ & rank-$5$ & rank-$10$ & rank-$20$ \\
  \hline
  \multirow{2}[0]{*}{Market-$1501$}
  &ours ($6$ parts)        & $\mathbf{80.4}$ & $\mathbf{91.5}$  & $\mathbf{94.3}$ & $\mathbf{96.4}$ \\
  &part seg. ($6$ parts)  & $61.2$          & $80.3$           &$86.9$          & $91.0$ \\
  \hline
  \multirow{2}[0]{*}{CUHK$03$}
  & ours ($6$ parts)  & $\mathbf{85.1}$ & $\mathbf{97.6}$ & $\mathbf{98.2}$ & $\mathbf{99.4}$ \\
  & part seg. ($6$ parts) & $70.7$ & $90.4$  & $94.8$ & $97.6$  \\
  \hline
\end{tabular}
}\vspace{-1.5em}
\end{table}

%

\vspace{0.1cm}
\noindent\textbf{Separate part segmentation.}
We conduct an experiment
with separate part segmentation.
We use the state-of-the-art part segmentation model~\cite{deeplabv1}
learnt from the PASCAL-Person-Part dataset~\cite{pascalpart} ($6$ part classes),
to compute the mask for both training and test images.
We modify our network
by replacing the masks from the part net
with the masks from the part segmentation model.
In the training stage,
we learn the modified network (the mask fixed)
using the same setting with our approach.

The results are shown in Table~\ref{tab:SeparatePartSegmentation}
and the performance is poor compared with our method.
This is reasonable
because the parts in our approach
are learnt directly for person re-identification
while the parts learnt from the PASCAL-Person-Part dataset
might not be very good
because it does not take consideration into the person re-identification problem.
We also think that
if the human part segmentation of the person re-identification training images
is available,
exploiting the segmentation as an extra supervision,
e.g., the learnt part corresponds to a human part,
or a sub-region of the human part,
is helpful for learning the part net.

\section{Experiments}


\subsection{Datasets}
%
\noindent\textbf{Market-$1501$}.
This dataset~\cite{market2015} is one of the largest benchmark datasets for person re-identification.
There are six cameras: $5$ high-resolution cameras,
and one low-resolution camera.
There are $32,668$ DPM-detected pedestrian image boxes of $1,501$ identities:
$750$ identifies are used for training
and the remaining $751$ for testing.
There are $3,368$ query images
and the large gallery (database) include $19,732$ images with $2,793$ distractors.

\vspace{0.1cm}
\noindent\textbf{CUHK$03$.}
This dataset~\cite{deepreid2014} consists of $13,164$ images of $1,360$ persons, captured by six cameras. Each identity only appears in two disjoint camera views, and there are on average $4.8$ images in each view. We use the provided training/test splits~\cite{deepreid2014} on the labeled data set.
For each test identity,
two images are randomly sampled
as the probe and gallery images, respectively,
and the average performance over $20$ trials
is reported as the final result.

\vspace{0.1cm}
\noindent\textbf{CUHK$01$.}
This dataset~\cite{cuhk2012} contains $971$ identities captured from two camera views in the same campus with CUHK$03$.
Each person has two images, each from one camera view.
Following the setup~\cite{improved2015},
we report the results of two different settings:
$100$ identifies for testing, and $486$ identities for testing.

\vspace{0.1cm}
\noindent\textbf{VIPeR.}
This dataset~\cite{viper2007} contains two views of $632$ persons.
Each pair of images about one person
are captured by different cameras with large viewpoint changes
and various illumination conditions.
The $632$ person images are divided into two halves,
$316$ for training and $316$ for testing.

\subsection{Evaluation Metrics}
We adopt the widely-used evaluation protocol~\cite{deepreid2014,improved2015}.
In the matching process,
we calculate the similarities
between each query and all the gallery images,
and then return the ranked list according to the similarities.
All the experiments are under the single query setting.
The performances are evaluated by
the cumulated matching characteristics (CMC) curves,
which is
an estimate of
the expectation of finding the correct match in
the top $n$ matches.
We also report the mean average precision (mAP) score~\cite{market2015}
over Market-$1501$.

\subsection{Empirical Analysis}\label{sec:empiricalstudy}
\noindent\textbf{The number of parts.}
We empirically study how the number of parts
affects the performance.
We conduct an experiment
over CUHK$03$:
randomly partition the training dataset
into two parts,
one for model learning
and the remaining for validation.
The performances for various numbers of parts,
$K=1, 2, 4, 8, 12$,
are given in Table~\ref{tab:NumberOfParts}.
It can be seen that
(\romannumeral1) more parts for the rank-$1$ score
lead to better scores
till $8$ parts
and then the scores become stable,
and (\romannumeral2) the scores of different number of parts at positions $5$, $10$, and $20$
are close
except the score of $1$ part at position $5$.
Thus, in our experiments,
we choose $K=8$ in the part net
for all the four datasets.
It is possible that
in other datasets
the optimal $K$ obtained through validation
is different.

\begin{table}
\centering
\caption{The~\emph{validation} performance with different numbers ($K$) of parts
over CUHK$03$.
The model is trained over a random half of the training data,
and the performance is reported
over the remaining half (as the validation set).
The best results are in bold.}
\label{tab:NumberOfParts}
\small{
\begin{tabular}{c|c c c c}
  \hline
  \#parts & rank-$1$ & rank-$5$ & rank-$10$ & rank-$20$ \\
  \hline
  $1$ & $77.7$   & $95.6$   & $98.4$ & $\mathbf{99.7}$ \\
  $2$ & $80.4$   & $96.7$   & $98.4$   & $99.4$\\
  $4$ & $82.0$   & $96.7$   & $\mathbf{98.8}$  & $\mathbf{99.7}$\\
  $8$ & $\mathbf{83.8}$  & $96.9$   & $98.3$   & $\mathbf{99.7}$\\
  $12$& $83.6$   & $\mathbf{97.3}$  & $\mathbf{98.8}$  & $99.6$\\
  \hline
\end{tabular}
}\vspace{-1.em}
\end{table}

\vspace{.1cm}
\noindent\textbf{Human segmentation and body part segmentation.}
The benefit from the body part segmentation lies in two points:
(\romannumeral1) remove the background
and (\romannumeral2) part alignment.
We compare our approach and the approach with human segmentation
that is implemented as our approach with $1$ part
and is able to remove the background.
The comparison shown from Table~\ref{tab:ComparisonWithHumanSegmentation}
over Market-$1501$ and CUHK$03$
shows that body part segmentation
performs superiorly in general.
The results imply that body part segmentation is beneficial.

\begin{table}
\centering
 \caption{
 The performances of our approach and human segmentation
 over Market-$1501$ and CUHK$03$.
 }
\label{tab:ComparisonWithHumanSegmentation}
\footnotesize{
\begin{tabular}{@{\hskip3pt}c@{\hskip3pt}|c|c c c |c}
  \hline
  Dataset & Method & rank-$1$ & rank-$5$ & rank-$10$ & mAP \\
  \hline
  \multirow{2}[0]{*}{Market-$1501$}
  &ours        & $\mathbf{81.0}$ & $\mathbf{92.0}$  & $\mathbf{94.7}$ & $\mathbf{63.4}$ \\
  &human seg.  & $74.2$          & $90.0$           & $93.8$          & $58.9$ \\
  \hline
  \multirow{2}[0]{*}{CUHK$03$}
  &ours        & $\mathbf{85.4}$ & $\mathbf{97.6}$  & $\mathbf{99.4}$ & $\mathbf{90.9}$\\
  &human seg. & $82.7$          & $95.9$           & $97.9$          & $88.6$ \\
  \hline
\end{tabular}
}\vspace{-1.5em}
\end{table}

\begin{table}
\centering
 \caption{
 The performances of our approach, two baseline networks without segmentation,
 modified by replacing the part net in our network with a fully-connected (FC) layer
 and an average pooling (pooling) layer
 over Market-$1501$
 and CUHK$03$.}
\label{tab:ComparisonWithBaseline}
\footnotesize{
\begin{tabular}{c|c|c c c | c}
  \hline
  Dataset & Method & rank-$1$ & rank-$5$ & rank-$10$ & mAP \\
  \hline
  \multirow{3}[0]{*}{Market-$1501$}
  & ours & $\mathbf{81.0}$ & $\mathbf{92.0}$ & $\mathbf{94.7}$ & $\mathbf{63.4}$ \\
  & FC & $75.9$ & $89.3$   & $92.9$   & $54.3$ \\
  & pooling & $75.9$ & $89.0$  & $92.2$ & $55.6$ \\
  \hline
  \multirow{3}[0]{*}{CUHK$03$}
  & ours & $\mathbf{85.4}$ & $\mathbf{97.6}$ & $\mathbf{99.4}$ & $\mathbf{90.9}$ \\
  & FC & $80.3$ & $95.5$ & $98.6$ & $87.3$ \\
  & pooling & $82.4$ & $96.8$ & $99.0$ & $88.9$ \\
  \hline
\end{tabular}
}
\vspace{-1.5em}
\end{table}

\vspace{.1cm}
\noindent\textbf{Comparison with non-human/part-segmentation.}
We compare the performances
of two baseline networks without segmentation,
which are modified from our network:
(\romannumeral1) replace the part net with a fully-connected layer
outputting the feature vector with the same dimension ($512$-d)
and (\romannumeral2) replace the part net with an global average-pooling layer
which also produces a $512$-d feature vector.

The fully-connected layer followed by the last convolutional layer
in (\romannumeral1)
has some capability to differentiate different spatial regions to some degree
through the linear weights,
which are however the same for all images,
yielding limited ability of differentiation.
The average-pooling method in (\romannumeral2)
ignores the spatial information,
though it is robust to the translations.
In contrast,
our approach is also able to differentiate body regions
and the differentiation is adaptive to each input image for translation/pose invariance.

The comparison over two datasets,
Market-$1501$ and CUHK$03$,
is given in Table~\ref{tab:ComparisonWithBaseline}.
It can be seen that
our approach outperforms these two baseline methods,
which indicates that
the part segmentation is capable of avoiding the mismatch due to part misalignment
in spatial partition
and improving the performance.

\begin{figure}[t]
\centering
    \includegraphics[width=.8\linewidth]{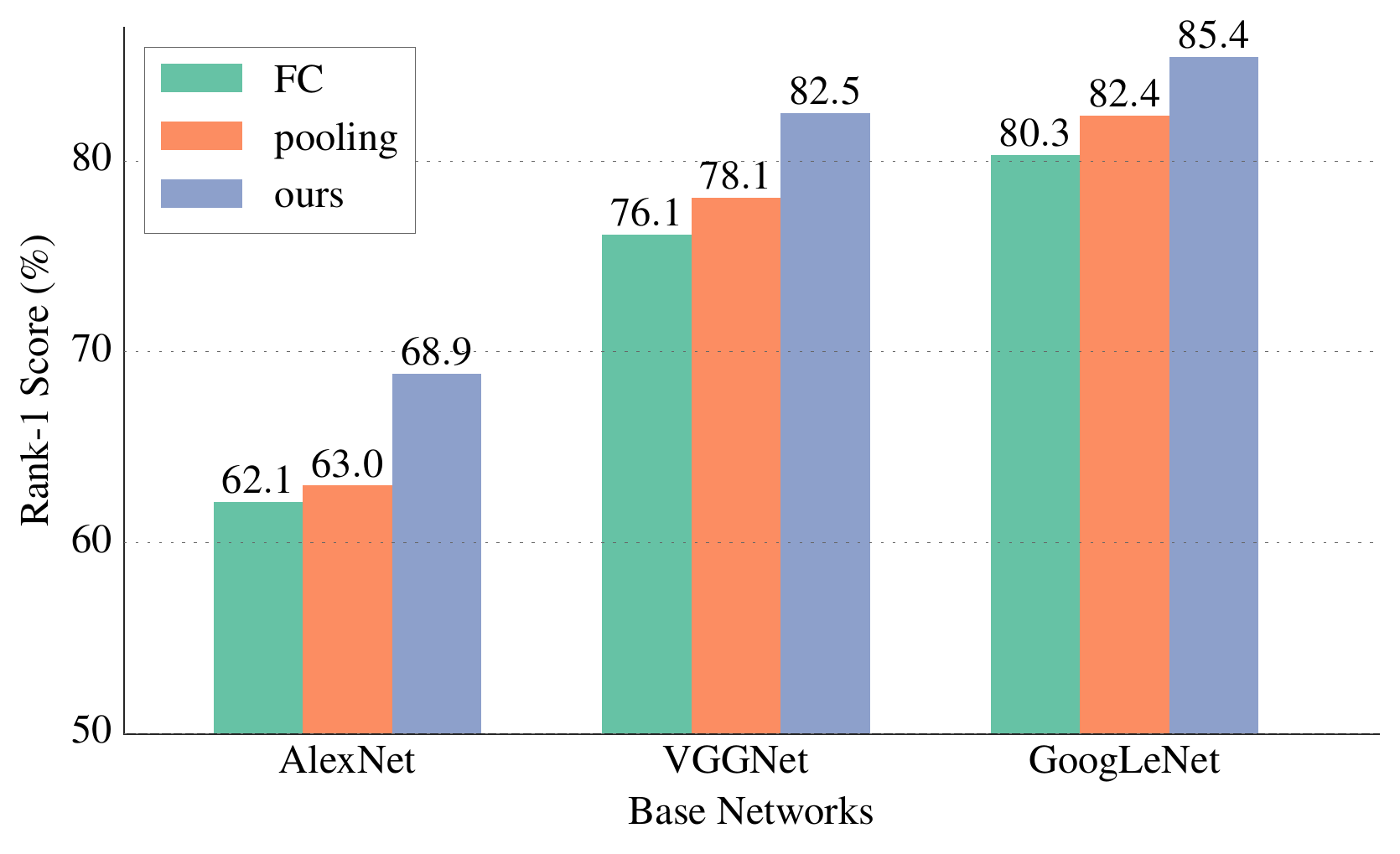}
\caption{The performance of our approach and the two baseline networks (FC and Pooling)
 with different feature map extraction networks over CUHK$03$.
 Our approach consistently boosts the performance
 for all the three networks (best viewed in color).}
\label{fig:resultswithdifferentnetworks}
\vspace{-1.em}
\end{figure}

\vspace{0.1cm}
\noindent\textbf{Image feature map extraction networks.}
We show that the part net can boost the performance
for various feature map extraction FCNs.
We report two extra results
with using AlexNet~\cite{alexnet2012} and VGGNet~\cite{vggnet2014}
as well as the result using GoogLeNet~\cite{googlenet2015}.
For AlexNet and VGGNet,
we remove the fully connected layers and use all the remaining convolutional layers
as the feature map extraction network,
and the training settings are the same as provided in Section~\ref{sec:implementationdetails}.
The results are depicted in Figure~\ref{fig:resultswithdifferentnetworks}.
It can be seen that
our approach consistently gets the performance gain for AlexNet, VGGNet and GoogLeNet.
In particular,
the gains with AlexNet and VGGNet are more significant:
compared with the baseline method with FC,
the gains are $6.8$, $6.4$, and $5.1$
for AlexNet, VGGNet and GoogLeNet, respectively,
and compared with the baseline method with pooling,
the gains are $5.9$, $4.4$, and $3.0$, respectively.

\vspace{.1cm}
\noindent\textbf{Comparison with other attention models.}
The part map detector is inspired by
the spatial attention model.
It is slightly different from
the standard attention model:
using sigmoid to replace softmax,
which brings more than $2\%$ gain
for rank-1 scores.
The comparative attention network (CAN) approach~\cite{LiuFQJY16}
is also based on the attention model
and adopts LSTM to help learn part maps.
It is not easy for us to have a good implementation
for CAN.
Thus, we report the results with AlexNet,
which CAN is based on,
as our base network.
The comparison is given in Table~\ref{tab:attentioncomparison}.
We can see that the overall performance of
our approach is better
except on the CUHK01 dataset for $100$ test IDs.

\begin{table}[t]
\footnotesize
\caption{\small{Compared with softmax over spatial responses and CAN~\cite{LiuFQJY16}.
All are based on AlexNet.
Larger is better.}}
\label{tab:attentioncomparison}
\resizebox{1\linewidth}{!}{
\begin{tabular}{c|c|c|c|c|c}
\hline
    \multicolumn{2}{c|}{} & rank-1 & rank-5 & rank-10 & rank-20\\
  \hline
  \multirow{3}{*}{CUHK03 (labeled)}
  & CAN & 65.65 &	91.28	& \textbf{96.29}	& 98.17 \\
  & \textbf{Ours} & \textbf{68.90}	& \textbf{91.40}	& 95.25	& \textbf{98.3} \\
  & Softmax & 65.14 & 90.64	& 95.43	& 97.79 \\
  \hline
  \multirow{3}{*}{CUHK03 (detected)}
  & CAN & 63.05 &	82.94 &	88.17	& 93.29 \\
  & \textbf{Ours} & \textbf{65.64} & \textbf{89.50} & \textbf{93.93} & \textbf{96.71}\\
  & Softmax & 64.36  & 89.50 & 94.71 & 97.43\\
  \hline
  \multirow{3}{*}{CUHK01-100}
  & CAN     & \textbf{81.04} & \textbf{96.89} & \textbf{99.67} & \textbf{100} \\
  & \textbf{Ours}    & 79.25 & 94.00 & 96.37 & 98.75 \\
  & Softmax & 74.64 & 91.27 & 94.55 & 97.27 \\
  \hline
  \multirow{3}{*}{Market}
  & CAN & 48.24 &	\multicolumn{3}{c}{mAP = 24.43} \\
  & \textbf{Ours} &  \textbf{64.22} &	\multicolumn{3}{c}{mAP = \textbf{41.80}} \\
  & Softmax & 62.23 &	\multicolumn{3}{c}{mAP = 41.01} \\
  \hline
\end{tabular}}
\vspace{-1.5em}
\end{table}

\subsection{Comparison with State-of-the-Arts}

\noindent\textbf{Market-$1501$.}
We compare our method with recent state-of-the-arts,
which are separated into four categories:
\underline{f}eature extraction (F),
\underline{m}etric learning (M),
\underline{d}eeply learnt feature representation (DF),
\underline{d}eep learning with \underline{m}atching sub\underline{n}etwork (DMN).
The results in Table~\ref{tab:results_market} are
obtained under the single query setting.

The competitive algorithm, pose-invariant embedding (PIE)~\cite{ZhengHLY17}
extracts part-aligned representation,
based on state-of-the-art pose estimator CPM~\cite{WeiRKS16} for part detection that is different from ours.
PIE uses ResNet-$50$ which is more powerful than GoogLeNet our approach uses.
We observe that our approach performs the best
and outperforms PIE:
$2.35$ gain for rank-$1$
and $9.5$ gain for mAP compared to PIE w/o using KISSME,
and $1.67$ for rank-$1$
and $7.4$ gain for mAP compared to PIE w/ using KISSME.

\begin{table}[t]
\centering
\caption{Performance comparison of state-of-the-art methods
on the recently released challenging dataset,
Market-$1501$. The methods are separated into four categories:
\underline{f}eature extraction (F),
\underline{m}etric learning (M),
\underline{d}eeply learnt feature representation (DF),
\underline{d}eep learning with \underline{m}atching sub\underline{n}etwork (DMN).}
\label{tab:results_market}%
\resizebox{1\linewidth}{!}
{
\begin{tabular}{c|l|cccc}
    \hline
    \multicolumn{2}{c|}{Method} & rank-1 & rank-5 & rank-10 & mAP \\
    \hline
    \multirow{2}[2]{*}{F} & LOMO~\cite{LOMO2015} (CVPR15) & $26.1$   & -     & -     & $7.8$  \\
    & BoW~\cite{market2015} (ICCV15) & $35.8$   & $52.4$   & $60.3$   & $14.8$  \\
    \hline
    & KISSME~\cite{kissme} (CVPR12) & $44.4$   & $63.9$   & $72.2$   & $20.8$  \\
    & WARCA~\cite{WARCA2016ECCV} (ECCV16) & $45.2$   & $68.2$   & $76.0$   & - \\
    M     & TMA~\cite{MartinelDMR16} (ECCV16) & $47.9$   & -     & -     & $22.3$  \\
    & SCSP~\cite{ChenYCZ16} (CVPR16) & $51.9$   & $72.0$   & $79.0$   & $26.4$  \\
    & DNS~\cite{null2016} (CVPR16) & $55.4$   & -     & -     & $29.9$  \\
    \hline
    \multirow{2}[1]{*}{DMN}
    & PersonNet~\cite{PersonNet2016} (ArXiv16) & $37.2$   & -     & -     & $18.6$  \\
    & Gated S-CNN~\cite{VariorHW16} (ECCV16) & $65.9$   & -     & -     & $39.6$  \\
    \hline
    \multirow{4}[1]{*}{DF}
    &  PIE~\cite{ZhengHLY17} & $78.65$ & $90.26$ & $93.59$ &  $53.87$  \\
          & PIE~\cite{ZhengHLY17} + KISSME~\cite{kissme} (Arxiv 2016) & $79.33$ & $90.76$ & $94.41$ & $55.95$ \\
         & SSDAL~\cite{SuZX0T16} (ECCV16) & $39.4$   & -     & -     & $19.6$  \\
         & Our Method & $\mathbf{81.0}$  & $\mathbf{92.0}$  & $\mathbf{94.7}$  & $\mathbf{63.4}$  \\
    \hline
\end{tabular}%
}
\vspace{-1em}
\end{table}%

\vspace{0.1cm}
\noindent\textbf{CUHK$03$.}
There are two versions of person boxes:
one is manually labeled
and the other one is detected
with a pedestrian detector.
We report the results for both versions
and all the previous results on CUHK$03$ are reported on the labeled version.
The results are given in Table~\ref{tab:results_cuhk03} for manually-labeled boxes
and in Table~\ref{tab:results_cuhk03_detected} for detected boxes.

Our approach performs the best on both versions.
On the one hand, the improvement over the detected boxes
is more significant
than that over the manually-labeled boxes.
This is because
the person body parts in the manually-labeled boxes
are spatially distributed more similarly.
On the other hand,
the performance of our approach over the manually-labeled boxes
are better than that over the detected-labeled boxes.
This means that the person position in the box (manually-labeled boxes are often better)
influences the part extraction quality,
which suggests that it is a necessity to learn a more robust part extractor
with more supervision information or over a larger dataset.

Compared with the competitive method DCSL~\cite{DCSL2016ijcai}
which is also based on the GoogLeNet,
the overall performance of our approach,
as shown in Table~\ref{tab:results_cuhk03},
is better on CUHK$03$
except that the rank-$5$ score of DCSL is slightly better
by $0.1\%$.
This is an evidence
demonstrating the powerfulness of the part-aligned representation
though DCSL adopts the strong matching subnetwork to improve the matching quality.
Compared with the second best method, PIE,
on the detected case as shown in Table~\ref{tab:results_cuhk03_detected},
our approach achieves $4.5$ gain at rank-$1$.

\begin{table}[t]
\centering
\caption{Performance comparison on CUHK$03$ for manually labeled human boxes. }
\label{tab:results_cuhk03}%
\resizebox{1\linewidth}{!}{
  \begin{tabular}{c|l|cccc}
    \hline
    \multicolumn{2}{c|}{Method} & rank-1 & rank-5 & rank-10 & rank-20 \\
    \hline
    & BoW~\cite{market2015} (ICCV15) & $18.9$   & $36.2$   & $46.8$   & - \\
    F   & LOMO~\cite{LOMO2015} (CVPR15) & $52.2$   & $82.2$   & $92.1$   & $96.3$  \\
    & GOG~\cite{MatsukawaOSS16} (CVPR16) & $67.3$   & $91.0$   & $96.0$   & - \\
    \hline
    & KISSME~\cite{kissme} (CVPR12) & $47.9$   & $69.3$   & $78.9$   & $87.0$  \\
    & SSSVM~\cite{ZhangLLIR16} (CVPR16) & $57.0$   & $84.8$   & $92.5$   & $96.4$  \\
    M   & DNS~\cite{null2016} (CVPR16) & $58.9$   & $85.6$   & $92.5$   & $96.3$  \\
    & Ensembles~\cite{PaisitkriangkraiSV15} (CVPR15) & $62.1$   & $89.1$   & $94.3$   & $97.8$  \\
    & WARCA~\cite{WARCA2016ECCV} (ECCV16) & $78.4$   & $94.6$   & $97.5$   & $99.1$  \\
    \hline
     \multirow{4}[1]{*}{DMN}
    & DeepReID~\cite{deepreid2014} (CVPR14) & $20.7$   & $51.3$   & $68.7$   & $83.1$  \\
    & IDLA~\cite{improved2015} (CVPR15) & $54.7$   & $86.4$   & $93.9$   & $98.1$  \\
    & PersonNet~\cite{PersonNet2016} (ArXiv16) & $64.8$   & $89.4$   & $94.9$   & $98.2$  \\
    & DCSL~\cite{DCSL2016ijcai} (IJCAI16) & $80.2$   & $\mathbf{97.7}$  & $99.2$   & $99.8$  \\
    \hline
    \multirow{2}[1]{*}{DF}
       & Deep Metric~\cite{ShiYZLLZL16} (ECCV16) & $61.3$   & $88.5$   & $96.0$   & $99.0$  \\
       & Our Method & $\mathbf{85.4}$  & $\mathbf{97.6}$  & $\mathbf{99.4}$  & $\mathbf{99.9}$  \\
    \hline
  \end{tabular}%
}\vspace{-1.em}
\end{table}%

\begin{table}[t]
\centering
\caption{Performance comparison on CUHK$03$ for detected boxes.}
\label{tab:results_cuhk03_detected}%
\resizebox{1\linewidth}{!}{
  \begin{tabular}{c|l|cccc}
    \hline
    \multicolumn{2}{c|}{Method} & rank-1 & rank-5 & rank-10 & rank-20 \\
    \hline
    \multirow{2}[2]{*}{F}
    & LOMO~\cite{LOMO2015} (CVPR15) & $46.3$   & $78.9$   & $88.6$   & $94.3$  \\
    & GOG~\cite{MatsukawaOSS16} (CVPR16) & $65.5$   & $88.4$   & $93.7$   & - \\
    \hline
    \multirow{4}[1]{*}{M}
    & LMNN~\cite{WeinbergerBS05} (NIPS05) & $6.3$    & $17.5$   & $28.2$   & $45.0$  \\
    & KISSME~\cite{kissme} (CVPR12) & $11.7$   & $33.9$   & $48.2$   & $65.0$  \\
    & SSSVM~\cite{ZhangLLIR16} (CVPR16) & $51.2$   & $81.5$   & $89.9$   & $95.0$  \\
    & DNS~\cite{null2016} (CVPR16) & $53.7$   & $83.1$   & $93.0$   & $94.8$  \\
    \hline
    \multirow{3}[1]{*}{DMN}
    & DeepReID~\cite{deepreid2014} (CVPR14) & $19.9$   & $50.0$   & $64.0$   & $78.5$  \\
    & IDLA~\cite{improved2015} (CVPR15) & $45.0$   & $76.0$   & $83.5$   & $93.2$  \\
    & SIR-CIR~\cite{WangZLZZ16} (CVPR16) & $52.2$   & $85.0$   & $92.0$   & $97.0$  \\
    \hline
    \multirow{3}[1]{*}{DF}
    & PIE~\cite{ZhengHLY17} + KISSME~\cite{kissme} (Arxiv 2016) & $67.10$ & $92.20$ & $96.60$ & $98.10$ \\
    & Deep Metric~\cite{ShiYZLLZL16} (ECCV16) & $52.1$   & $84.0$   & $92.0$   & $96.8$  \\
    & Our Method & $\mathbf{81.6}$  & $\mathbf{97.3}$  & $\mathbf{98.4}$  & $\mathbf{99.5}$  \\
    \hline
  \end{tabular}%
}\vspace{-1.em}
\end{table}%

\vspace{0.1cm}
\noindent\textbf{CUHK$01$.}
There are two evaluation settings~\cite{improved2015}:
$100$ test IDs, and $486$ test IDs.
Since there are a small number ($485$) of training identities for the case of $486$ test IDs,
as done in~\cite{improved2015,chen2016deep,DCSL2016ijcai},
we fine-tune the model, which is learnt
from the CUHK$03$ training set,
over the $485$ training identities:
the rank-$1$ score from the model learnt from CUHK$03$ is $44.59\%$
and it becomes $72.3\%$ with the fine-tuned model.

The results are reported in Table~\ref{tab:results_cuhk01_100}
and Table~\ref{tab:results_cuhk01_486}, respectively.
Our approach performs the best among the algorithms
w/o using matching subnetwork.
Compared to the competitive algorithm DCSL ~\cite{DCSL2016ijcai}
that uses matching subnetwork,
we can see that for $100$ test IDs,
our approach performs better in general
except a slightly low rank-$1$ score
and that for $486$ test IDs
our initial approach performs worse
and with a simple trick, removing one pooling layer
to double the feature map size,
the performance is much closer.
One notable point is that our approach is
advantageous in scaling up to large datasets.


\begin{table}[t]
\centering
\caption{Performance comparison
on CUHK$01$ for $100$ test IDs.}
\label{tab:results_cuhk01_100}%
\resizebox{1\linewidth}{!}{
\begin{tabular}{c|l|cccc}
    \hline
    \multicolumn{2}{c|}{Method} & rank-1 & rank-5 & rank-10 & rank-20 \\
    \hline
    \multirow{6}[1]{*}{DMN}
        & DeepReID~\cite{deepreid2014} (CVPR14) & $27.9$   & $58.2$   & $73.5$   & $86.3$  \\
        & IDLA~\cite{improved2015} (CVPR15) & $65.0$   & $88.7$   & $93.1$   & $97.2$  \\
        & Deep Ranking (~\cite{chen2016deep} (TIP16)) & $50.4$   & $70.0$   & $84.8$   & $92.0$  \\
        & PersonNet~\cite{PersonNet2016} (ArXiv16) & $71.1$   & $90.1$   & $95.0$   & $98.1$  \\
        & SIR-CIR~\cite{WangZLZZ16} (CVPR16) & $71.8$   & $91.6$   & $96.0$   & $98.0$  \\
        & DCSL~\cite{DCSL2016ijcai} (IJCAI16) & $\mathbf{89.6}$  & $97.8$   & $98.9$   & $99.7$  \\
    \hline
    \multirow{2}[1]{*}{DF}
        & Deep Metric~\cite{ShiYZLLZL16} (ECCV16) & $69.4$   & $90.8$   & $96.0$   & - \\
       & Our Method & $\mathbf{88.5}$  & $\mathbf{98.4}$  & $\mathbf{99.6}$  & $\mathbf{99.9}$  \\
    \hline
\end{tabular}%
}
\vspace{-1.em}
\end{table}%

\begin{table}[t]
\centering
\caption{Performance comparison
on CUHK$01$ for $486$ test IDs.}
\label{tab:results_cuhk01_486}%
\resizebox{1\linewidth}{!}{
\begin{tabular}{c|l|cccc}
    \hline
    \multicolumn{2}{c|}{Method} & rank-1 & rank-5 & rank-10 & rank-20 \\
    \hline
    \multirow{4}[2]{*}{F}
    & Semantic~\cite{ShiHX15} (CVPR15) & $31.5$   & $52.5$   & $65.8$   & $77.6$  \\
    & MirrorRep~\cite{ChenZL15} (IJCAI15) & $40.4$   & $64.6$   & $75.3$   & $84.1$  \\
    & LOMO~\cite{LOMO2015} (CVPR15) & $49.2$   & $75.7$   & $84.2$   & $90.8$  \\
    & GOG~\cite{MatsukawaOSS16} (CVPR16) & $57.8$   & $79.1$   & $86.2$   & $92.1$  \\
    \hline
    & LMNN~\cite{WeinbergerBS05} (NIPS05) & $13.5$   & $31.3$   & $42.3$   & $54.1$  \\
    & SalMatch~\cite{salience2013iccv} (ICCV13) & $28.5$   & $45.9$   & $55.7$   & $68.0$  \\
  M & DNS~\cite{null2016} (CVPR16) & $65.0$   & $85.0$   & $89.9$   & $94.4$  \\
    & WARCA~\cite{WARCA2016ECCV} (ECCV16) & $65.6$   & $85.3$   & $90.5$   & $95.0$  \\
    & SSSVM~\cite{ZhangLLIR16} (CVPR16) & $66.0$   & $89.1$   & $92.8$   & $96.5$  \\
     \hline \multirow{3}[2]{*}{DMN}
    & IDLA~\cite{improved2015} (CVPR15) & $47.5$   & $71.6$   & $80.3$   & $87.5$  \\
   & Deep Ranking~\cite{chen2016deep} (TIP16) & $50.4$   & $70.0$   & $84.8$   & $92.0$  \\
   & DCSL~\cite{DCSL2016ijcai} (IJCAI16) & $\mathbf{76.5}$ & $\mathbf{94.2}$ &  $\mathbf{97.5}$ & - \\
    \hline
    \multirow{2}[2]{*}{DF}
    & TCP-CNN~\cite{ChengGZWZ16} (CVPR16) & $53.7$   & $84.3$   & $91.0$   & $96.3$  \\
    & Our Method & ${72.3}$  & ${91.0}$  & ${94.9}$  & ${97.2}$  \\
    & Our Method + remove pool3 & $\mathbf{75.0}$ & $\mathbf{93.5}$ & $\mathbf{95.7}$ & $\mathbf{97.7}$  \\
    \hline
\end{tabular}%
}\vspace{-1.em}
\end{table}%


\begin{table}[t]
\centering
\caption{Results on a relatively small dataset, VIPeR.
}
\label{tab:results_viper}%
\resizebox{1\linewidth}{!}{
\begin{tabular}{c|l|cccc}
    \hline
    \multicolumn{2}{c|}{Method} & rank-1 & rank-5 & rank-10 & rank-20\\
    \hline
    \multirow{6}[1]{*}{F}
    & ELF~\cite{viewpoint2008} (ECCV 2008) & $12.0$  & $44.0$  & $47.0$  & $61.0$ \\
    & BoW~\cite{market2015} (ICCV15) & $21.7$   & $42.0$   & $50.0$   & $60.9$ \\
    & LOMO~\cite{LOMO2015} (CVPR15) & $40.0$   & $68.1$   & $80.5$   & $91.1$  \\
    & Semantic~\cite{ShiHX15} (CVPR15) & $41.6$   & $71.9$   & $86.2$   & $95.1$  \\
    & MirrorRep~\cite{ChenZL15} (IJCAI15) & $43.0$   & $75.8$   & $87.3$   & $94.8$  \\
    & GOG~\cite{MatsukawaOSS16} (CVPR16) & $\mathbf{49.7}$  & $\mathbf{79.7}$  & $\mathbf{88.7}$  & $\mathbf{94.5}$ \\
    \hline
    \multirow{8}[1]{*}{M}    & LMNN~\cite{WeinbergerBS05} (NIPS05) & $11.2$   & $32.3$   & $44.8$   & $59.3$ \\
    & KISSME~\cite{kissme} (CVPR12) & $19.6$   & $47.5$   & $62.2$   & $77.0$  \\
    & LADF~\cite{LiCLHCS13} (CVPR13) & $30.0$   & $64.7$   & $79.0$   & $91.3$  \\
    & WARCA~\cite{WARCA2016ECCV} (ECCV16) & $37.5$   & $70.8$   & $82.0$   & $92.0$  \\
    & DNS~\cite{null2016} (CVPR16) & $42.3$   & $71.5$   & $82.9$   & $92.1$  \\
    & SSSVM~\cite{ZhangLLIR16} (CVPR16) & $42.7$   & -     & $84.3$   & $91.9$  \\
    & TMA~\cite{MartinelDMR16} (ECCV16) & $43.8$   & -     & $83.9$   & $91.5$  \\
    & SCSP~\cite{ChenYCZ16} (CVPR16) & $\mathbf{53.5}$  & $\mathbf{82.6}$  & $\mathbf{91.5}$  & $\mathbf{96.7}$ \\
    \hline
    \multirow{6}[1]{*}{DMN}
    & IDLA~\cite{improved2015} (CVPR15) & $34.8$   & $63.6$   & $75.6$   & $84.5$ \\
    & Gated S-CNN~\cite{VariorHW16} (ECCV16) & $37.8$   & $66.9$   & $77.4$   & - \\
    & SSDAL~\cite{SuZX0T16} (ECCV16) & $37.9$   & $65.5$   & $75.6$   & $88.4$  \\
    & SIR-CIR~\cite{WangZLZZ16} (CVPR16) & $35.8$   & $67.4$   & $83.5$   & - \\
    & Deep Ranking~\cite{chen2016deep} (TIP16) & ${38.4}$   & ${69.2}$   & ${81.3}$   & ${90.4}$  \\
    & DCSL~\cite{DCSL2016ijcai} (IJCAI16) & $\mathbf{44.6}$   & $\mathbf{73.4}$   & $\mathbf{82.6}$   & $\mathbf{91.9}$  \\
    \hline
    \multirow{5}[1]{*}{DF}
    & PIE~\cite{ZhengHLY17} + Mirror~\cite{ChenZL15} + MFA~\cite{YanXZZYL07} (Arxiv 2016) & $43.3$ & $69.4$ & $80.4$ & $90.0$ \\
    & Fusion~\cite{ZhengHLY17} + MFA~\cite{YanXZZYL07} (Arxiv 2016) &  $\mathbf{54.5 }$ &  $\mathbf{84.4}$ & $\mathbf{92.2}$ & $\mathbf{96.9}$ \\
    & Deep Metric~\cite{ShiYZLLZL16} (ECCV16) & $40.9$   & $67.5$   & $79.8$   & - \\
    & TCP-CNN~\cite{ChengGZWZ16} (CVPR16) & $47.8$   & ${74.7}$  & $84.8$   & $91.1$ \\
    & Our Method & ${48.7}$  & ${74.7}$  & ${85.1}$  & ${93.0}$ \\
    \hline
\end{tabular}%

}\vspace{-1.em}
\end{table}%

\vspace{0.1cm}
\noindent\textbf{VIPeR.}
The dataset is relatively small and the training images are not enough for training.
We fine-tune the model learnt from CUHK$03$ following~\cite{VariorHW16,improved2015}.
The results are presented in Table~\ref{tab:results_viper}.
Our approach outperforms other deep learning-based approaches
except PIE~\cite{ZhengHLY17} with complicated schemes
while performs poorer than the best-performed feature extraction approach GOG~\cite{MatsukawaOSS16}
and metric learning method SCSP~\cite{ChenYCZ16}.
In comparison with PIE~\cite{ZhengHLY17},
our approach performs better than PIE with data augmentation
Mirror~\cite{ChenZL15} and metric learning MFA~\cite{YanXZZYL07}
and lower than PIE with a more complicated fusion scheme,
which our approach might benefit from.
In general,
the results suggest that
like in other tasks, e.g., classification,
training deep neural networks from a small data
is still an open and challenging problem.

\vspace{0.1cm}
\noindent\textbf{Summary.}
The overall performance of our approach
is the best in the category of deeply-learnt feature representation (DF)
and better than non-deep learning algorithms
except in the small dataset VIPeR.
In comparison to the category
of deep learning with matching subnetwork (DMN),
our approach in general is good,
and performs worse than DCSL in CUHK$01$ with $486$ test IDs.
It is reasonable as matching network is more complicated
than the simple Euclidean distance in our approach.
One notable advantage is that our approach
is efficient in online matching and cheap in storage,
while DCSL stores large feature maps of gallery images for
online similarity computation, resulting in
larger storage cost and higher
online computation cost.

\section{Conclusions}
In this paper,
we present a novel part-aligned representation approach
to handle the body misalignment problem.
Our formulation follows the idea of attention models
and is in a deep neural network form,
which is learnt only from person similarities
without the supervision information about the human parts.
Our approach aims to partition the human body
instead of the human image box into grids or strips,
and thus is more robust to pose changes
and different human spatial distributions in the human image box
and thus the matching is more reliable.
Our approach learns more useful body parts for person re-identification
than separate body part detection.

\section*{Acknowledgements}
This work was
supported in part by 
the National Natural Science Foundation of China
under Grant U1509206 and Grant 61472353, in part by the
National Basic Research Program of China under Grant Grant 2015CB352302,
and partially funded by the MOE-Microsoft Key Laboratory of 
Visual Perception, Zhejiang University.

{\small
\bibliographystyle{ieee}
\bibliography{references}

\begin{thebibliography}{10}\itemsep=-1pt

\bibitem{improved2015}
E.~Ahmed, M.~Jones, and T.~K. Marks.
\newblock An improved deep learning architecture for person re-identification.
\newblock In {\em CVPR}, 2015.

\bibitem{BakCBT10a}
S.~Bak, E.~Corv{\'{e}}e, F.~Br{\'{e}}mond, and M.~Thonnat.
\newblock Person re-identification using spatial covariance regions of human
  body parts.
\newblock In {\em AVSS}, pages 435--440, 2010.

\bibitem{ChenYCZ16}
D.~Chen, Z.~Yuan, B.~Chen, and N.~Zheng.
\newblock Similarity learning with spatial constraints for person
  re-identification.
\newblock In {\em CVPR}, June 2016.

\bibitem{dapeng2015}
D.~Chen, Z.~Yuan, G.~Hua, N.~Zheng, and J.~Wang.
\newblock Similarity learning on an explicit polynomial kernel feature map for
  person re-identification.
\newblock In {\em CVPR}, pages 1565--1573, 2015.

\bibitem{deeplabv1}
L.-C. Chen, G.~Papandreou, I.~Kokkinos, K.~Murphy, and A.~L. Yuille.
\newblock Semantic image segmentation with deep convolutional nets and fully
  connected crfs.
\newblock In {\em ICLR}, 2015.

\bibitem{chen2016deep}
S.-Z. Chen, C.-C. Guo, and J.-H. Lai.
\newblock Deep ranking for person re-identification via joint representation
  learning.
\newblock {\em {IEEE} Trans. Image Processing}, 25(5):2353--2367, 2016.

\bibitem{pascalpart}
X.~Chen, R.~Mottaghi, X.~Liu, S.~Fidler, R.~Urtasun, and A.~Yuille.
\newblock Detect what you can: Detecting and representing objects using
  holistic models and body parts.
\newblock In {\em CVPR}, pages 1971--1978, 2014.

\bibitem{ChenZL15}
Y.~Chen, W.~Zheng, and J.~Lai.
\newblock Mirror representation for modeling view-specific transform in person
  re-identification.
\newblock In {\em IJCAI}, pages 3402--3408, 2015.

\bibitem{ChengGZWZ16}
D.~Cheng, Y.~Gong, S.~Zhou, J.~Wang, and N.~Zheng.
\newblock Person re-identification by multi-channel parts-based cnn with
  improved triplet loss function.
\newblock In {\em CVPR}, June 2016.

\bibitem{ChengCSBM11}
D.~S. Cheng, M.~Cristani, M.~Stoppa, L.~Bazzani, and V.~Murino.
\newblock Custom pictorial structures for re-identification.
\newblock In {\em BMVC}, pages 1--11, 2011.

\bibitem{tripletdeepreid2015}
S.~Ding, L.~Lin, G.~Wang, and H.~Chao.
\newblock Deep feature learning with relative distance comparison for person
  re-identification.
\newblock {\em Pattern Recognition}, 48(10):2993--3003, 2015.

\bibitem{Farenzena2010symmetry}
M.~Farenzena, L.~Bazzani, A.~Perina, V.~Murino, and M.~Cristani.
\newblock Person re-identification by symmetry-driven accumulation of local
  features.
\newblock In {\em CVPR}, pages 2360--2367, June 2010.

\bibitem{GarciaMMG15}
J.~Garcia, N.~Martinel, C.~Micheloni, and A.~Gardel.
\newblock Person re-identification ranking optimisation by discriminant context
  information analysis.
\newblock In {\em ICCV}, December 2015.

\bibitem{viper2007}
D.~Gray, S.~Brennan, and H.~Tao.
\newblock Evaluating appearance models for recognition, reacquisition, and
  tracking.
\newblock In {\em Proc. IEEE International Workshop on PETS}, 2007.

\bibitem{viewpoint2008}
D.~Gray and H.~Tao.
\newblock Viewpoint invariant pedestrian recognition with an ensemble of
  localized features.
\newblock In {\em ECCV}, 2008.

\bibitem{jia2014caffe}
Y.~Jia, E.~Shelhamer, J.~Donahue, S.~Karayev, J.~Long, R.~B. Girshick,
  S.~Guadarrama, and T.~Darrell.
\newblock Caffe: Convolutional architecture for fast feature embedding.
\newblock {\em CoRR}, abs/1408.5093, 2014.

\bibitem{JingZWYKYHX15}
X.-Y. Jing, X.~Zhu, F.~Wu, X.~You, Q.~Liu, D.~Yue, R.~Hu, and B.~Xu.
\newblock Super-resolution person re-identification with semi-coupled low-rank
  discriminant dictionary learning.
\newblock In {\em CVPR}, June 2015.

\bibitem{WARCA2016ECCV}
C.~Jose and F.~Fleuret.
\newblock Scalable metric learning via weighted approximate rank component
  analysis.
\newblock In {\em ECCV}, 2016.

\bibitem{KodirovXFG16}
E.~Kodirov, T.~Xiang, Z.~Fu, and S.~Gong.
\newblock Person re-identification by unsupervised $\ell_1$ graph learning.
\newblock In {\em ECCV}, pages 178--195, 2016.

\bibitem{kissme}
M.~K{\"o}stinger, M.~Hirzer, P.~Wohlhart, P.~M. Roth, and H.~Bischof.
\newblock Large scale metric learning from equivalence constraints.
\newblock In {\em CVPR}, pages 2288--2295, 2012.

\bibitem{alexnet2012}
A.~Krizhevsky, I.~Sutskever, and G.~E. Hinton.
\newblock Imagenet classification with deep convolutional neural networks.
\newblock In {\em NIPS}, pages 1106--1114, 2012.

\bibitem{cuhk2012}
W.~Li, R.~Zhao, and X.~Wang.
\newblock Human reidentification with transferred metric learning.
\newblock In {\em ACCV}, 2012.

\bibitem{deepreid2014}
W.~Li, R.~Zhao, T.~Xiao, and X.~Wang.
\newblock Deepreid: Deep filter pairing neural network for person
  re-identification.
\newblock In {\em CVPR}, 2014.

\bibitem{LiZWXG15}
X.~Li, W.-S. Zheng, X.~Wang, T.~Xiang, and S.~Gong.
\newblock Multi-scale learning for low-resolution person re-identification.
\newblock In {\em ICCV}, December 2015.

\bibitem{LiCLHCS13}
Z.~Li, S.~Chang, F.~Liang, T.~S. Huang, L.~Cao, and J.~R. Smith.
\newblock Learning locally-adaptive decision functions for person verification.
\newblock In {\em CVPR}, pages 3610--3617, 2013.

\bibitem{LOMO2015}
S.~Liao, Y.~Hu, X.~Zhu, and S.~Z. Li.
\newblock Person re-identification by local maximal occurrence representation
  and metric learning.
\newblock In {\em CVPR}, 2015.

\bibitem{psd2015}
S.~Liao and S.~Z. Li.
\newblock Efficient psd constrained asymmetric metric learning for person
  re-identification.
\newblock In {\em ICCV}, 2015.

\bibitem{LiuFQJY16}
H.~Liu, J.~Feng, M.~Qi, J.~Jiang, and S.~Yan.
\newblock End-to-end comparative attention networks for person
  re-identification.
\newblock {\em CoRR}, abs/1606.04404, 2016.

\bibitem{ma2012local}
B.~Ma, Y.~Su, and F.~Jurie.
\newblock Local descriptors encoded by fisher vectors for person
  re-identification.
\newblock In {\em ECCV}, pages 413--422, 2012.

\bibitem{MartinelDMR16}
N.~Martinel, A.~Das, C.~Micheloni, and A.~K. Roy{-}Chowdhury.
\newblock Temporal model adaptation for person re-identification.
\newblock In {\em ECCV}, pages 858--877, 2016.

\bibitem{MatsukawaOSS16}
T.~Matsukawa, T.~Okabe, E.~Suzuki, and Y.~Sato.
\newblock Hierarchical gaussian descriptor for person re-identification.
\newblock In {\em CVPR}, June 2016.

\bibitem{mignon2012pcca}
A.~Mignon and F.~Jurie.
\newblock Pcca: A new approach for distance learning from sparse pairwise
  constraints.
\newblock In {\em CVPR}, pages 2666--2672, 2012.

\bibitem{PaisitkriangkraiSV15}
S.~Paisitkriangkrai, C.~Shen, and A.~van~den Hengel.
\newblock Learning to rank in person re-identification with metric ensembles.
\newblock In {\em CVPR}, June 2015.

\bibitem{PengXWPGHT16}
P.~Peng, T.~Xiang, Y.~Wang, M.~Pontil, S.~Gong, T.~Huang, and Y.~Tian.
\newblock Unsupervised cross-dataset transfer learning for person
  re-identification.
\newblock In {\em CVPR}, June 2016.

\bibitem{facenet2015}
F.~Schroff, D.~Kalenichenko, and J.~Philbin.
\newblock Facenet: A unified embedding for face recognition and clustering.
\newblock In {\em CVPR}, 2015.

\bibitem{correspondence2015}
Y.~Shen, W.~Lin, J.~Yan, M.~Xu, J.~Wu, and J.~Wang.
\newblock Person re-identification with correspondence structure learning.
\newblock In {\em ICCV}, 2015.

\bibitem{ShiYZLLZL16}
H.~Shi, Y.~Yang, X.~Zhu, S.~Liao, Z.~Lei, W.~Zheng, and S.~Z. Li.
\newblock Embedding deep metric for person re-identification: {A} study against
  large variations.
\newblock In {\em ECCV}, pages 732--748, 2016.

\bibitem{ShiHX15}
Z.~Shi, T.~M. Hospedales, and T.~Xiang.
\newblock Transferring a semantic representation for person re-identification
  and search.
\newblock In {\em CVPR}, June 2015.

\bibitem{vggnet2014}
K.~Simonyan and A.~Zisserman.
\newblock Very deep convolutional networks for large-scale image recognition.
\newblock {\em CoRR}, abs/1409.1556, 2014.

\bibitem{SuYZTDLG15}
C.~Su, F.~Yang, S.~Zhang, Q.~Tian, L.~S. Davis, and W.~Gao.
\newblock Multi-task learning with low rank attribute embedding for person
  re-identification.
\newblock In {\em ICCV}, December 2015.

\bibitem{SuZX0T16}
C.~Su, S.~Zhang, J.~Xing, W.~Gao, and Q.~Tian.
\newblock Deep attributes driven multi-camera person re-identification.
\newblock In {\em ECCV}, pages 475--491, 2016.

\bibitem{googlenet2015}
C.~Szegedy, W.~Liu, Y.~Jia, P.~Sermanet, S.~Reed, D.~Anguelov, D.~Erhan,
  V.~Vanhoucke, and A.~Rabinovich.
\newblock Going deeper with convolutions.
\newblock In {\em CVPR}, 2015.

\bibitem{VariorHW16}
R.~R. Varior, M.~Haloi, and G.~Wang.
\newblock Gated siamese convolutional neural network architecture for human
  re-identification.
\newblock In {\em ECCV}, pages 791--808, 2016.

\bibitem{VariorSLXW16}
R.~R. Varior, B.~Shuai, J.~Lu, D.~Xu, and G.~Wang.
\newblock A siamese long short-term memory architecture for human
  re-identification.
\newblock In {\em ECCV}, pages 135--153, 2016.

\bibitem{WangZLZZ16}
F.~Wang, W.~Zuo, L.~Lin, D.~Zhang, and L.~Zhang.
\newblock Joint learning of single-image and cross-image representations for
  person re-identification.
\newblock In {\em CVPR}, June 2016.

\bibitem{WangGZX16}
H.~Wang, S.~Gong, X.~Zhu, and T.~Xiang.
\newblock Human-in-the-loop person re-identification.
\newblock In {\em ECCV}, pages 405--422, 2016.

\bibitem{WeiRKS16}
S.~Wei, V.~Ramakrishna, T.~Kanade, and Y.~Sheikh.
\newblock Convolutional pose machines.
\newblock In {\em {CVPR}}, pages 4724--4732, 2016.

\bibitem{WeinbergerBS05}
K.~Q. Weinberger, J.~Blitzer, and L.~K. Saul.
\newblock Distance metric learning for large margin nearest neighbor
  classification.
\newblock In {\em NIPS}, pages 1473--1480, 2005.

\bibitem{PersonNet2016}
L.~Wu, C.~Shen, and A.~van~den Hengel.
\newblock Personnet: Person re-identification with deep convolutional neural
  networks.
\newblock {\em CoRR}, abs/1601.07255, 2016.

\bibitem{wu2016enhanced}
S.~Wu, Y.-C. Chen, X.~Li, A.-C. Wu, J.-J. You, and W.-S. Zheng.
\newblock An enhanced deep feature representation for person re-identification.
\newblock In {\em WACV}, pages 1--8, 2016.

\bibitem{XiaoLOW16}
T.~Xiao, H.~Li, W.~Ouyang, and X.~Wang.
\newblock Learning deep feature representations with domain guided dropout for
  person re-identification.
\newblock In {\em CVPR}, June 2016.

\bibitem{xiaoli2016end}
T.~Xiao, S.~Li, B.~Wang, L.~Lin, and X.~Wang.
\newblock End-to-end deep learning for person search.
\newblock {\em CoRR}, abs/1604.01850, 2016.

\bibitem{XuBKCCSZB15}
K.~Xu, J.~Ba, R.~Kiros, K.~Cho, A.~C. Courville, R.~Salakhutdinov, R.~S. Zemel,
  and Y.~Bengio.
\newblock Show, attend and tell: Neural image caption generation with visual
  attention.
\newblock In {\em {ICML}}, pages 2048--2057, 2015.

\bibitem{XuLZL13}
Y.~Xu, L.~Lin, W.~Zheng, and X.~Liu.
\newblock Human re-identification by matching compositional template with
  cluster sampling.
\newblock In {\em ICCV}, pages 3152--3159, 2013.

\bibitem{YanXZZYL07}
S.~Yan, D.~Xu, B.~Zhang, H.~Zhang, Q.~Yang, and S.~Lin.
\newblock Graph embedding and extensions: {A} general framework for
  dimensionality reduction.
\newblock {\em {IEEE} Trans. Pattern Anal. Mach. Intell.}, 29(1):40--51, 2007.

\bibitem{deepmetric2014}
D.~Yi, Z.~Lei, S.~Liao, and S.~Z. Li.
\newblock Deep metric learning for person re-identification.
\newblock In {\em ICLR}, 2014.

\bibitem{null2016}
L.~Zhang, T.~Xiang, and S.~Gong.
\newblock Learning a discriminative null space for person re-identification.
\newblock In {\em CVPR}, 2016.

\bibitem{ZhangLLIR16}
Y.~Zhang, B.~Li, H.~Lu, A.~Irie, and X.~Ruan.
\newblock Sample-specific svm learning for person re-identification.
\newblock In {\em CVPR}, June 2016.

\bibitem{DCSL2016ijcai}
Y.~Zhang, X.~Li, L.~Zhao, and Z.~Zhang.
\newblock Semantics-aware deep correspondence structure learning for robust
  person re-identification.
\newblock In {\em IJCAI}, pages 3545--3551, 2016.

\bibitem{salience2013iccv}
R.~Zhao, W.~Ouyang, and X.~Wang.
\newblock Person re-identification by salience matching.
\newblock In {\em ICCV}, 2013.

\bibitem{ZhaoOW14}
R.~Zhao, W.~Ouyang, and X.~Wang.
\newblock Learning mid-level filters for person re-identification.
\newblock In {\em CVPR}, June 2014.

\bibitem{ZhengBSWSWT16}
L.~Zheng, Z.~Bie, Y.~Sun, J.~Wang, C.~Su, S.~Wang, and Q.~Tian.
\newblock {MARS:} {A} video benchmark for large-scale person re-identification.
\newblock In {\em ECCV}, pages 868--884, 2016.

\bibitem{ZhengHLY17}
L.~Zheng, Y.~Huang, H.~Lu, and Y.~Yang.
\newblock Pose invariant embedding for deep person re-identification.
\newblock {\em CoRR}, abs/1701.07732, 2017.

\bibitem{market2015}
L.~Zheng, L.~Shen, L.~Tian, S.~Wang, J.~Wang, and Q.~Tian.
\newblock Scalable person re-identification: A benchmark.
\newblock In {\em ICCV}, 2015.

\bibitem{ZhengWTHLT15}
L.~Zheng, S.~Wang, L.~Tian, F.~He, Z.~Liu, and Q.~Tian.
\newblock Query-adaptive late fusion for image search and person
  re-identification.
\newblock In {\em CVPR}, June 2015.

\bibitem{ZhengLXLLG15}
W.-S. Zheng, X.~Li, T.~Xiang, S.~Liao, J.~Lai, and S.~Gong.
\newblock Partial person re-identification.
\newblock In {\em ICCV}, December 2015.

\end{thebibliography}
}

\end{document}